\documentclass[11pt]{article}

\usepackage[final]{acl}

\usepackage{times}
\usepackage{latexsym}

\usepackage[T1]{fontenc}
\usepackage[utf8]{inputenc}

\usepackage{microtype}

\usepackage{inconsolata}

\usepackage{graphicx}
\graphicspath{{../}}

\makeatletter
\def\input@path{{../}}
\makeatother

\usepackage{tabularx}
\usepackage{multirow}
\usepackage{booktabs}
\usepackage{adjustbox}

\usepackage{algorithm}
\usepackage{algorithmic}
\usepackage{listings}

\usepackage{pgf}
\usepackage{tikz}
\usepackage{tcolorbox}
\tcbuselibrary{skins, breakable}

\usepackage{pifont}
\usepackage{dsfont}
\usepackage{mathtools}
\usepackage{amssymb}
\usepackage{colortbl}
\usepackage{arydshln}

\usepackage{enumitem}

\definecolor{mygreen}{HTML}{2ca02c}
\definecolor{myred}{HTML}{d62728}
\definecolor{myyellow}{HTML}{ff7f0e}
\definecolor{bestcolor}{HTML}{E6E0F8}
\definecolor{secondbestcolor}{HTML}{FAEBDD}
\definecolor{thirdbestcolor}{HTML}{E9E9E9}

\definecolor{brandblue}{rgb}{0.34, 0.7, 1}

\lstdefinestyle{promptstyle}{
    basicstyle=\ttfamily\small,
    breaklines=true,
    breakindent=0pt,
    showstringspaces=false,
    frame=none,
    numbers=none,
    keepspaces=true,
    columns=fullflexible,
    tabsize=2,
    escapeinside={(*}{*)},
    moredelim=**[is][\textcolor{blue}]{@}{@},
}

\tcbset{
  base/.style={
    arc=0mm,
    bottomtitle=-0.25mm,
    boxrule=0mm,
    colbacktitle=black!10!white,
    coltitle=black,
    fonttitle=\bfseries,
    left=2.5mm,
    leftrule=1mm,
    right=3.5mm,
    title={#1},
    toptitle=0.25mm,
    breakable,
  }
}
\newtcolorbox{mybox}[1]{
  colframe=brandblue,
  base={#1}
}

\newcommand{\yes}{\textcolor{mygreen}{\ding{51}}}
\newcommand{\no}{\textcolor{myred}{\ding{55}}}
\newcommand{\half}{\textcolor{myyellow}{\ding{52}\rotatebox[origin=c]{-9.2}{\kern-0.7em\ding{55}}}}

\title{Behavior2Trip: Towards Personalized Travel Planning \\via User Behavior Trajectory}

\author{%
Zihao Cheng\textsuperscript{1*},
Yingyu Shan\textsuperscript{2*},
Hongru Wang\textsuperscript{3*},
Zeming Liu\textsuperscript{1$\dagger$},
Xinyi Wang\textsuperscript{4},\\
\textbf{Xiangrong Zhu}\textsuperscript{4$\ddagger$},
\textbf{Yuhang Guo\textsuperscript{2},} 
\textbf{Wei Lin\textsuperscript{4},}
\textbf{Yunhong Wang\textsuperscript{1}} \\
\textsuperscript{1}School of Computer Science and Engineering, Beihang University, Beijing \\
\textsuperscript{2}Beijing Institute of Technology \quad
\textsuperscript{3}University of Edinburgh \quad 
\textsuperscript{4}Meituan Inc. \\
\textsuperscript{*}Equal Contribution \quad  \textsuperscript{$\dagger$}Corresponding Author  \quad \textsuperscript{$\ddagger$}Project Leader\\\
Email: \texttt{zihaocheng@buaa.edu.cn, zmliu@buaa.edu.cn}
}
\begin{document}
\maketitle

\begin{abstract}
Travel planning agents assist users in generating personalized travel plans by modeling their individual preferences. Existing agents either rely on explicit user instructions or engage in multi-turn clarification to elicit user preferences. However, both approaches overlook the rich behavioral signals latent in users' past behaviors, which implicitly encode their preferences. This over-reliance on active user input increases interaction burden and limits plan personalization. To bridge this gap, we introduce a new task, \textbf{Behavior-Aware Travel Planning}, which infers user preferences directly from past behaviors and generates personalized travel plans. To facilitate research on this task, we introduce \textbf{Behavior2Trip}, a benchmark constructed from one of the largest Chinese online travel platforms, comprising 11,400 instances. Each instance represents an average of 39.8 past user behaviors spanning 14 attributes across 5 preference dimensions. We further propose \textbf{B2T-Agent}, a reinforcement learning-based agent that leverages user behavior trajectories, interacts with external tools for preference-aligned retrieval, and maintains an internal memory module. Experiments on Behavior2Trip show that GPT-4.1 achieves a full-constraint pass rate of only 0.5\% on the hardest tasks, while B2T-Agent built upon Qwen3-8B outperforms all baselines, highlighting the substantial challenge of this task. Moreover, Qwen3-8B trained with B2T-Agent also outperforms GPT-4.1 on the TravelPlanner benchmark, demonstrating strong generalization\footnote{Code and data are available at \url{https://github.com/BUAA-IRIP-LLM/Behavior2Trip}.}.
\end{abstract}

\section{Introduction}
\begin{figure}[!ht]
    \centering
    \includegraphics[width=\linewidth]{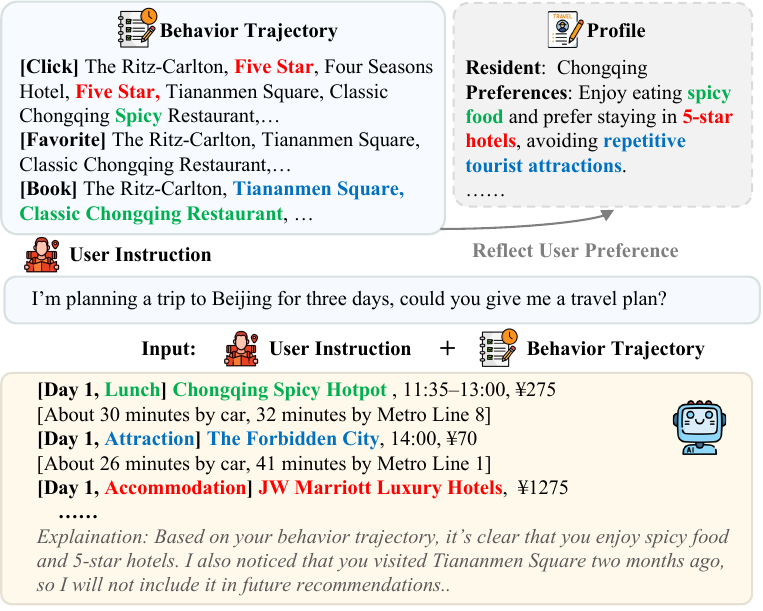}
    \caption{\textbf{An example of Behavior2Trip}, where the agent analyzes a user's behavioral trajectory to implicitly infer preferences, enabling the direct generation of personalized travel plans from implicit instructions.}
    \label{fig:intro}
\end{figure}

Travel planning agents have become an essential role for assisting users in complex travel decision-making \cite{xie2024travelplanner, shao2025chinatravelopenendedbenchmarklanguage, zhang2024ask, ni2025tpragbenchmarkingretrievalaugmentedlarge, retail}. With the advancement of Large Language Models (LLMs)~\citep{ma2025advancing, he2025llm2rec, cheng2025generative, mem2evolve, nc}, these agents are gaining the ability to interact with external environments to generate comprehensive travel plans \cite{jiang2024delegationdesigningidealagentic,ju2024globettglanguagedrivenguaranteed,luo2025largelanguagemodelagent,zhang2025agentmodelsinternalizingchainofaction,singh2025agenticreasoningtoolintegration,zhang2025nemotronresearchtooln1exploringtoolusinglanguage}, paving the way for more practical and intelligent travel solutions.

\definecolor{rowblue}{RGB}{235, 245, 255}
\begin{table*}[!ht]
\centering
\begin{adjustbox}{width=0.9\textwidth, center}
\begin{tabular}{lccccccccccc}
\toprule[0.08em]
\textbf{Dataset} & \textbf{UBT} & \textbf{SII} & \textbf{OP} & \textbf{UPF} & \textbf{LIB} & \textbf{UP} & \textbf{\# POIs} & \textbf{\# POI Attr.} & \textbf{\# Constraints} & \textbf{Instances} \\
\midrule[0.05em]
TravelPlanner \cite{xie2024travelplanner} & \no & \no & \yes & \no & \no & exp. & 19.9k & 6.6 & 15 & 20 \\
ChinaTravel \cite{shao2025chinatravelopenendedbenchmarklanguage} & \no & \no & \yes & \no & \no & exp. & 12.2k & 6.8 & 23 & 154 \\
Ask--before--plan \cite{zhang2024ask} & \no & \yes & \no & \no & \no & exp. & 19.9k & 6.6 & 12 & 2,000 \\
ITINERA \cite{tang2024itinera} & \no & \no & \yes & \no & \no & exp. & 7.6k & 9.0 & -- & -- \\
TRIP--PAL \cite{delarosa2024trippaltravelplanningguarantees} & \no & \no & \yes & \no & \no & exp. & 1.8k & -- & 2 & 100 \\
Travel--Agent \cite{chen2024travelagent} & \no & \no & \no & \no & \no & exp. & -- & -- & 20 & 20 \\
TP--RAG \cite{ni2025tpragbenchmarkingretrievalaugmentedlarge} & \no & \no & \yes & \no & \no & exp. & 85.6k & -- & -- & 2,348 \\
\midrule[0.05em]
\rowcolor{rowblue}
\textbf{Behavior2Trip (ours)} & \yes & \yes & \yes & \yes & \yes & \textbf{imp.} & \textbf{80.9k} & \textbf{17.7} & \textbf{32} & \textbf{11,400} \\
\bottomrule[0.08em]
\end{tabular}
\end{adjustbox}
\caption{\textbf{Comparison between the Behavior2Trip and other benchmarks.} \textbf{UBT}: User Behavior Trajectory, \textbf{SII}: Support Implicit Instruction, \textbf{OP}: One--shot Planning, \textbf{UPF}: User Profile, \textbf{LIB}: Low Interaction Burden, \textbf{UP}: User Preference. The \yes~indicates full support, the \no~indicates no support, ``--'' denotes not mentioned in paper.}
\label{tab:dataset_comparison}
\end{table*}

Existing travel planning agents follow two main paradigms: \textbf{(1) Single-turn Explicit Instruction} \cite{xie2024travelplanner,shao2025chinatravelopenendedbenchmarklanguage,ni2025tpragbenchmarkingretrievalaugmentedlarge,delarosa2024trippaltravelplanningguarantees}, which requires users to provide all preferences upfront in a single turn; and \textbf{(2) Multi-turn Clarification} \cite{zhang2024ask,chen2024travelagent}, which interactively elicits preferences through dialogue. However, both paradigms heavily rely on users' active input, overlooking behavioral signals from users' past interactions (e.g., clicks, favorites, and bookings) that naturally encode their preferences \cite{ye2011exploiting, chen2023survey}. This over-reliance on active input increases interaction burden and limits plan personalization.

To bridge this gap, we introduce a new task, \textbf{Behavior-Aware Travel Planning}, which infers user preferences directly from past behaviors and generates personalized travel plans without requiring active user input. To support this task, we construct \textbf{Behavior2Trip}, a benchmark grounded in real user data from one of the largest Chinese online travel platforms, comprising 11,400 instances, each with an average of 39.8 past user behaviors spanning 14 attributes across 5 preference dimensions. The dataset is built upon a sandbox environment with 80.6k real-world POIs, where behavior trajectories are generated from diverse preference profiles and validated through a rigorous two-stage quality control process.

To address the challenge of implicitly inferring user preferences from behavior trajectories while satisfying long-horizon planning constraints, we propose \textbf{B2T-Agent}, a reinforcement learning-based agent that generates personalized travel plans from user behavior trajectories. Unlike prompt-driven methods \cite{yuan-etal-2025-evoagent, xie2024travelplanner}, B2T-Agent features a structured action space supporting both \textbf{external tool invocation} and \textbf{internal memory management}, optimized by a composite reward function that rewards plan quality and penalizes invalid actions.

We conduct comprehensive experiments on Behavior2Trip to evaluate both open-sourced and closed-sourced models. Results highlight the substantial challenge of this task: even GPT-4.1 achieves a full-constraint pass rate of only 0.5\% on the hardest tasks, while B2T-Agent built upon Qwen3-8B significantly outperforms all baselines. Moreover, Qwen3-8B trained with B2T-Agent also outperforms GPT-4.1 on the TravelPlanner~\citep{xie2024travelplanner}, demonstrating strong generalization. Overall, the contributions of this paper are as follows:
\begin{itemize}[leftmargin=*]
    \item We identify a novel task, \textbf{Behavior-Aware Travel Planning}, which generates personalized travel plans by inferring user preferences directly from past behaviors, without requiring explicit or iterative user input.

    \item To support this task, we introduce \textbf{Behavior2Trip}, a large-scale benchmark grounded in real user data, comprising 11,400 instances with rich behavioral signals spanning 14 attributes across 5 preference dimensions.

    \item To tackle the challenges of this task, we propose \textbf{B2T-Agent}, a reinforcement learning-based agent with structured external tool invocation and internal memory management. Experiments show that B2T-Agent built upon Qwen3-8B significantly outperforms GPT-4.1 and generalizes well to the TravelPlanner benchmark.

\end{itemize}

\section{Related Work}

\begin{figure*}[!ht]
    \centering
    \includegraphics[width=\textwidth]{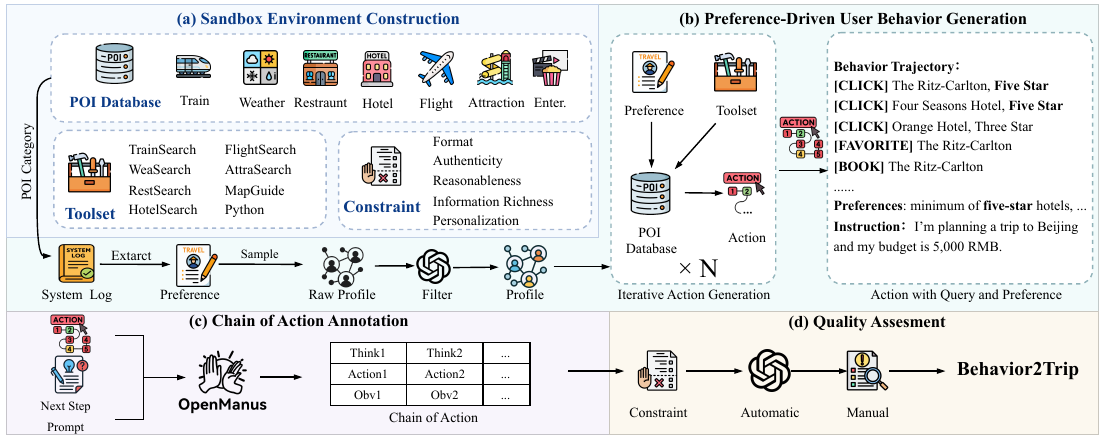}
    \caption{\textbf{The overall construction pipeline of Behavior2Trip}, comprising four phases: (a) \textbf{Sandbox Environment Construction} builds a realistic platform with real-world POIs, tools, and constraints; (b) \textbf{User Behavior Generation} produces user trajectories across three difficulty levels; (c) \textbf{Chain of Action Annotation} collects agent trajectories with chain-of-action reasoning paths; and (d) \textbf{Quality Assessment} ensures data quality through two-stage automated and human verification.}
    \label{fig:data-collection}
\end{figure*}

\subsection{Travel Planning Benchmarks}
Travel planning is a representative task for evaluating agents' long-horizon planning capabilities, as it requires understanding user instructions, performing multi-turn tool use to gather information, and generating constraint-satisfying travel plans. Existing benchmarks fall into two paradigms: \textbf{(1) Single-turn Explicit Instruction} \cite{xie2024travelplanner, shao2024chinatravel, de2024trip, singh2024personal}, where the agent generates a complete travel plan from a single, explicit user request; and \textbf{(2) Multi-turn Clarification} \cite{zhang2024ask, chen2024travelagent}, where the agent interactively elicits user preferences through dialogue to refine the plan. However, as shown in Table~\ref{tab:dataset_comparison}, both paradigms require users to actively express their preferences and cannot address scenarios where users express their preferences implicitly. In contrast, \textbf{Behavior2Trip} evaluates agents' ability to infer user preferences from behavior trajectories and generate preference-aligned travel plans, enabling more realistic evaluation of personalized planning in real-world scenarios.

\subsection{Travel Planning Agents}
To build travel planning agents, existing methods generally fall into two paradigms. \textbf{(1) End-to-End LLM Agent} \cite{xie2024travelplanner, chen2024travelagent, zhang2024ask, tang2024itinera, liu2026towards, peap}, which relies on manually designed prompts to guide information retrieval and plan generation. However, such handcrafted workflows lack flexibility and struggle to generalize across diverse user inputs. \textbf{(2) Hybrid LLM-Symbolic Solver} \cite{hao2024large, ju2024globe, de2024trip, shao2024chinatravel}, which improves logical consistency by incorporating symbolic solvers, but typically applies them only at the final planning stage, leaving the step-by-step information gathering process unguided. To address these limitations, we propose \textbf{B2T-Agent}, a reinforcement learning-based agent that eliminates handcrafted components and learns to interact with external tools and manage internal memory throughout the entire planning process.
\section{Behavior2Trip}

\subsection{Problem Definition} \label{sec:definition}
We define the user's action trajectory as \( \mathcal{A} = [a_1, a_2, \dots, a_t] \), where each \( a_i \) denotes an action (e.g., click, favorite, book) at time step $i$, implicitly encoding the user's travel preferences \(\mathcal{P}\). Given a natural-language travel request \( \mathcal{I} \), the model queries the POI database \( \mathcal{D} \) via toolset \( \mathcal{T} \) to generate a personalized travel plan \(\mathcal{R} = \mathrm{Model}(\mathcal{A}, \mathcal{I}, \mathcal{T}, \mathcal{D})\), subject to two constraints: \ding{172} \textbf{Commonsense Constraint (CC)} (\( \mathcal{R} \in \mathcal{C}_{CC} \)): the plan must be logically feasible and factually grounded; \ding{173} \textbf{User Preference Constraint (UPC)} (\( \mathcal{R} \in \mathcal{C}_{UPC} \)): the plan must align with the user's inferred preferences.

\subsection{Data Collection} \label{sec:data collection}

As shown in Figure \ref{fig:data-collection}, this section describes the step-by-step process of constructing a diverse and realistic dataset, which includes four key stages: (a) Sandbox Environment Construction, (b) Preference-Driven User Behavior Generation, (c) Chain of Action Annotation, and (d) Quality Assessment.

\subsubsection{Sandbox Environment Construction} \label{sec: sandbox}
To create a realistic and fully functional sandbox environment, we establish a \textit{POI Database}, a \textit{Toolset}, and \textit{Constraints} sourced from one of the largest Chinese online travel platforms. All data is fully anonymized with no user privacy information involved.

\paragraph{POI Database} We collect 80.6k real POIs from 24 popular Chinese travel cities, including \textit{Attraction}, \textit{Hotel}, \textit{Restaurant}, and \textit{Entertainment}. Each POI has an average of 17.7 fields, including: \ding{172} basic info (e.g., name, category), \ding{173} spatiotemporal data (e.g., location, hours, pricing), \ding{174} user feedback (e.g., reviews, ratings). We also include 728.4k records on \textit{Flight}, \textit{Train}, and \textit{Weather}, forming a comprehensive POI database $\mathcal{D}$. See Appendix~\ref{app:poi} for details.

\paragraph{Toolset} We design a fine-grained toolset where each POI category has a dedicated search tool supporting optional filtering fields such as location, price range, rating, and opening hours. We also implement \textit{MapGuide}, which helps interpret spatial data, and \textit{Python} for general-purpose computing. In total, we develop 9 tools; details are in Appendix~\ref{app:toolset}.

\paragraph{Constraints} To enable fine-grained evaluation of travel plans while preserving real-world fidelity, we design a constraint set grounded in authentic online travel service scenarios. For \textbf{CC}, we define four dimensions: \textit{Format} ensures structural consistency, \textit{Authenticity} verifies POI-related correctness, \textit{Reasonableness} checks logical feasibility, and \textit{Information Richness} reflects plan completeness. For \textbf{UPC}, we introduce the \textit{Personalization} dimension to capture alignment with user preferences. Altogether, we define 32 constraints; details are in Appendix~\ref{app:constraint}.

\subsubsection{User Behavior Generation}
To construct user trajectories that are both preference-grounded and behaviorally coherent, we first derive a structured preference schema from real logs, instantiate user profiles with controlled difficulty, and then use LLM role-playing to generate long-term click--favorite--book sequences.

\definecolor{rowblue}{RGB}{235, 245, 255}
\begin{table}[t]
\centering
\begin{adjustbox}{width=0.9\columnwidth, center}
\begin{tabular}{l|ccc|ccc}
\toprule[0.08em]
\multirow{2}{*}{\textbf{Statistic}} & \multicolumn{3}{c|}{\textbf{Train}} & \multicolumn{3}{c}{\textbf{Test}} \\
\cmidrule(lr){2-4} \cmidrule(lr){5-7}
& \textbf{E} & \textbf{M} & \textbf{H} & \textbf{E} & \textbf{M} & \textbf{H} \\
\midrule[0.05em]
\rowcolor{rowblue}
\multicolumn{7}{c}{\textit{\textbf{Dataset}}} \\
% \cdashline{1-7}
Samples      &      3000&      3000&      3000&      800&      800&      800\\
\midrule[0.05em]
\rowcolor{rowblue}
\multicolumn{7}{c}{\textit{\textbf{Trajectory}}} \\
% \cdashline{1-7}
Actions      &      14.5&      38.2&      66.7&      14.2&      37.6& 67.2      \\
Preferences  &      1.0&      2.4&      4.0&      1.0&      2.3& 4.0       \\
POIs         &      10.3&      27.2&      47.8&      10.1&      26.9& 48.0      \\
\midrule[0.05em]
\rowcolor{rowblue}
\multicolumn{7}{c}{\textit{\textbf{Instruction}}} \\
% \cdashline{1-7}
Trip Duration       &      4.9&      4.9&      5.0&      4.9&      4.9& 5.0       \\
Companions   &      2.2&      2.1&      2.3&      2.2&      2.2& 2.2       \\
\bottomrule[0.08em]
\end{tabular}
\end{adjustbox}
\vspace{-3mm}
\caption{\textbf{Statistics of Behavior2Trip.  } E, M, and H represent Easy, Medium, and Hard levels, respectively.}
\label{tab:dataset_statistics}
\vspace{-3mm}
\end{table}

\paragraph{Preference Schema} 
We derive a structured preference schema from anonymized user logs by extracting \textbf{14} high-frequency attributes across \textbf{5} dimensions. For each attribute, we calibrate its value range and POI-category mapping against real-log distributions, with details in Appendix~\ref{app:preference}.

\paragraph{User Profile} 
We instantiate user profiles $\mathcal{P}$ at three difficulty levels to control preference complexity and behavioral noise. \textit{Easy} includes one preference category without noise; \textit{Medium} combines multiple categories with one noise behavior; and \textit{Hard} covers all categories with a higher noise ratio. For each level, we sample attribute combinations from the schema and use GPT-4.1 to remove inconsistent profiles, with the prompt shown in Figure~\ref{prompt-profile}.

\paragraph{Behavior Trajectory} 
Motivated by prior studies showing that user check-in trajectories are shaped by sequential, spatio-temporal, and semantic/category patterns~\cite{cheng2013you,tian2023next,wang2022point}, we model real user sessions as coherent travel-intent trajectories. To capture this structure, for each profile $\mathcal{P}$, we construct a candidate pool containing preference-matched POIs and a difficulty-controlled set of noise POIs. Gemini-3-Flash is then prompted to role-play the user and generate coherent \textbf{click} $\rightarrow$ \textbf{add\_to\_favorites} $\rightarrow$ \textbf{book} sequences (prompt shown in Figure~\ref{prompt-trajectory}). Noise POIs are restricted to click actions to simulate exploratory browsing, while favorite and book actions remain preference-aligned. Repeating this process across sessions yields long-term trajectories $\mathcal{A}$, which are paired with real travel instructions $\mathcal{I}$ from system logs to form $(\mathcal{A}, \mathcal{P}, \mathcal{I})$ triplets.

\begin{figure*}[!ht]
    \centering
    \includegraphics[width=\textwidth]{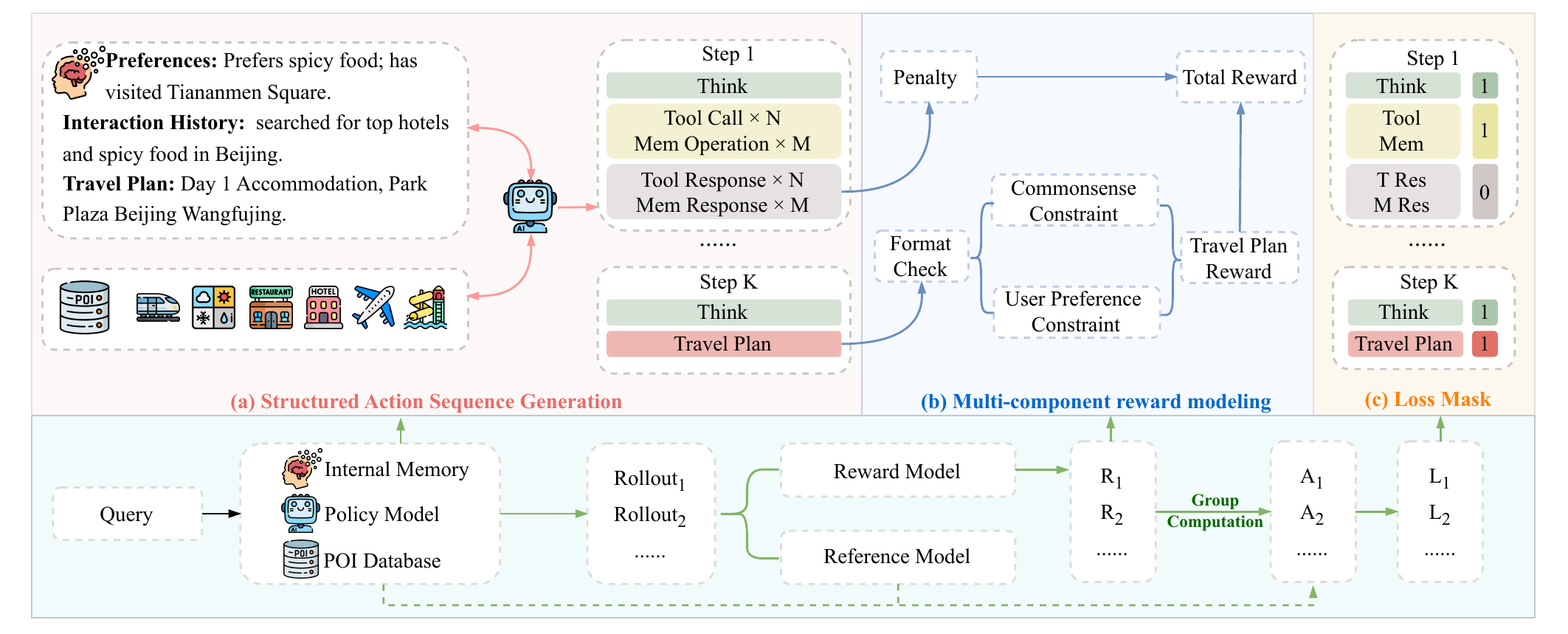}
    \caption{\textbf{Overview of the proposed B2T-Agent framework.} B2T-Agent incorporates \textbf{external tool invocation} and \textbf{internal memory management}, optimized via Group Relative Policy Optimization with a multi-component reward function that rewards plan quality and penalizes invalid actions.}
    \label{fig:method}
\end{figure*}

\subsubsection{Chain of Action Annotation} 
After obtaining the $(\mathcal{A}, \mathcal{P}, \mathcal{I})$ triplets, we annotate each instance with a \textbf{Chain of Action (CoA)} label that records the full reasoning trajectory---including tool calls and intermediate decisions---to provide step-level supervision for model training. To generate these trajectories, we use OpenManus~\cite{openmanus2025} as the rollout harness, feeding $\mathcal{A}$ and $\mathcal{I}$ as inputs. Since unconstrained rollouts tend to drift from the correct planning direction, after each \texttt{(think, action, observation)} cycle we manually inject a tailored \textbf{Next Step Prompt} (Figure~\ref{prompt-next-step-prompt}) to steer the agent toward the next correct step, ensuring the resulting trajectory remains coherent and usable as a supervision signal.

\subsubsection{Quality Assessment}
To ensure dataset quality, we use a two-stage filtering process~\citep{durecdail, durecdail2, terminalworld}. In the first stage, we utilized Deepseek-R1 to automatically score each instance on a 0-3 scale with respect to three aspects: trajectory completeness, reflection of user preferences, and constraint satisfaction (Figure~\ref{prompt-quality-control}). Only instances that received a score of 3 were retained. Subsequently, three independent annotators reviewed a random sample of 1000 instances, reporting no observable errors and achieving 94.6\% agreement.

\subsection{Data Statistics}

Table~\ref{tab:dataset_statistics} summarizes dataset statistics across three difficulty levels. The full dataset comprises 11,400 instances (9,000 train / 2,400 test). With increasing difficulty, the number of actions, preference categories, and involved POIs all grow, making it progressively harder for models to infer user preferences from behavioral trajectories and generate constraint-satisfying plans. At each difficulty level, instructions span varied trip durations and group sizes, ensuring the benchmark captures a broad distribution of real-world travel planning scenarios.
\section{B2T-Agent}
In this section, we present B2T-Agent, an RL-based agent that supports external tool invocation and internal memory management for personalized travel planning. As shown in Figure~\ref{fig:method}, B2T-Agent comprises three key components: (a) Structured Action Sequence Generation (\S\ref{sec:rollout}), (b) Multi-Component Reward Modeling (\S\ref{sec:reward}), and (c) Training with Loss Mask (\S\ref{sec:lossmask}).

\subsection{Structured Action Sequence Generation} \label{sec:rollout}
To support iterative POI retrieval from the database and coherent context management across long-horizon planning, we define the agent's rollout as a structured action sequence $y$:
\begin{equation*}
    y \coloneq (a_1, a_2, \dots, a_k, R)
\end{equation*}
where each action $a_i$ is one of the following types: External Tool Use $\mathcal{T}$, Internal Memory Management $\mathcal{M}$, or Chain-of-Thought Reasoning $\mathcal{C}$. The sequence concludes with a final Result $\mathcal{R}$.

\paragraph{External Tool Use}
The agent wraps each tool invocation in \texttt{<tool\_call>} and receives the result in \texttt{<tool\_response>}. Built on Qwen-Agent and following the Model Context Protocol\footnote{\url{https://www.anthropic.com/news/model-context-protocol}}, it can access any tool in our sandbox. To improve efficiency, multiple tools can be called within a single reasoning step.

\paragraph{Internal Memory Management}
To mitigate context window saturation from accumulating trajectories and tool responses, the agent maintains an internal memory module organized as key-value pairs, accessible via \texttt{<memory>}. The model can \textbf{read} entries by key and \textbf{write} new entries to store user preferences, tool outputs, and interaction history across planning steps.

\paragraph{Reasoning and Result}
To infer user preferences from behavior trajectories and guide subsequent tool invocations, the model performs explicit reasoning within \texttt{<think>} to analyze retrieved information and intermediate results. Once all necessary interactions are complete, it generates the final travel plan within \texttt{<answer>}.

\subsection{Multi-Component Reward Modeling} \label{sec:reward}
Given the structured action sequence defined above, we design a multi-component reward function to supervise both the process of tool invocation and the quality of the final travel plan, comprising an Incorrect Action Penalty and a Travel Plan Reward.

\definecolor{rowblue}{RGB}{235, 245, 255}
\begin{table*}[!ht]
    \centering
    \renewcommand{\arraystretch}{1.35}
    \begin{adjustbox}{max width=\textwidth, scale=1}
    \begin{tabular}{ll*{3}{ccccc}c}
        \toprule[0.08em]
        % --- 表头 ---
        \multirow{2}{*}{\textbf{Model}} & \multirow{2}{*}{\textbf{Method}} & \multicolumn{5}{c}{\textbf{Easy}} & \multicolumn{5}{c}{\textbf{Medium}} & \multicolumn{5}{c}{\textbf{Hard}} & \multirow{2}{*}{\textbf{Avg.}} \\
        \cmidrule(lr){3-7} \cmidrule(lr){8-12} \cmidrule(lr){13-17}
        & & \textbf{DR} & \textbf{CC} & \textbf{UPC} & \textbf{Final} & \textbf{LLM} & \textbf{DR} & \textbf{CC} & \textbf{UPC} & \textbf{Final} & \textbf{LLM} & \textbf{DR} & \textbf{CC} & \textbf{UPC} & \textbf{Final} & \textbf{LLM} & \\
        \midrule
        
        % --- ReAct 数据 ---
        Qwen3-8B    & \multirow{6}{*}{ReAct} & 40.2 & 34.2/0.0 & 10.1/10.1 & 0.0 & 27.5 & 34.1 & 31.8/0.0 & 12.3/0.0 & 0.0 & 23.5 & 11.2 & 24.1/0.0 & 15.9/0.0 & 0.0 & 12.8 & 0.0 \\
        Qwen3-14B   &                        & 44.3 & 32.9/0.3 & 14.2/14.2 & 0.3 & 32.6 & 32.8 & 32.9/0.4 & 19.7/1.8 & 1.8 & 28.7 & 21.5 & 31.9/1.2 & 21.1/0.0 & 0.0 & 23.2 & 0.7 \\
        Qwen3-32B   &                        & 53.1 & 32.4/0.3 & 14.3/14.3 & 0.3 & 34.3 & 41.5 & 32.5/0.8 & 19.8/0.8 & 0.8 & 29.9 & 31.4 & 32.4/1.3 & 22.7/0.0 & 0.0 & 33.3 & 0.4 \\
        GPT-4o      &                        & 90.4 & 36.5/1.3 & 16.8/16.8 & 1.3 & 36.6 & \underline{97.7} & 36.4/4.2 & 19.4/3.1 & 3.1 & 31.9 & 96.2 & 33.1/3.4 & 22.2/0.3 & 0.3 & 26.9 & 1.5 \\
        GPT-4.1     &                        & 95.8 & 33.6/2.2 & 17.7/17.7 & 2.2 & 36.4 & 95.0 & 33.4/3.0 & 28.4/7.0 & \underline{7.0} & 31.2 & \textbf{97.8} & 33.5/3.1 & 22.2/0.5 & 0.5 & 28.2 & 3.2 \\
        DeepSeek-V3 &                        & 94.3 & 34.7/3.3 & 20.1/20.1 & 3.1 & 41.6 & 94.9 & 33.9/3.7 & 28.4/4.4 & 4.4 & 38.8 & 97.0 & 34.1/4.1 & 31.8/0.8 & 0.5 & 34.1 & 2.7 \\
        \midrule

        % --- SFT 数据 ---
        Qwen3-8B    & \multirow{2}{*}{SFT}   & 77.2 & 54.0/5.0 & 17.7/17.7 & 5.0 & 73.7 & 76.5 & 54.4/4.6 & 26.5/3.7 & 3.7 & 62.3 & 57.4 & 54.5/4.1 & 32.9/0.3 & 0.3 & 35.8 & 3.0 \\
        Qwen3-14B   &                        & 87.3 & \underline{58.7}/5.2 & 17.8/17.8 & \underline{5.2} & 73.1 & 88.6 & \underline{59.0}/\underline{5.5} & 29.0/4.1 & 4.1 & 66.9 & 61.5 & \underline{55.2}/5.0 & 33.2/0.3 & 0.3 & 36.1 & 3.2 \\
        \midrule

        % --- B2T-Agent 数据 ---
        \rowcolor{rowblue}
        Qwen3-8B    &   & \textbf{100.0} & 56.8/\underline{5.2} & \underline{25.0}/\underline{25.0} & 4.7 & \underline{77.2} & 96.3 & 57.3/5.2 & \underline{37.2}/\underline{10.3} & 5.2 & \underline{70.2} & 93.3 & 53.3/\underline{5.1} & \underline{36.9}/\underline{4.3} & \underline{4.3} & \underline{68.5} & \underline{4.7} \\
        \rowcolor{rowblue}
        Qwen3-14B   & \multirow{-2}{*}{B2T-Agent}  & \underline{98.9} & \textbf{59.8}/\textbf{7.3} & \textbf{27.9}/\textbf{27.9} & \textbf{7.3} & \textbf{81.1} & \textbf{99.0} & \textbf{59.3}/\textbf{6.2} & \textbf{40.3}/\textbf{13.5} & \textbf{6.2} & \textbf{75.2} & \underline{97.4} & \textbf{59.4}/\textbf{6.1} & \textbf{41.2}/\textbf{5.0} & \textbf{5.0} & \textbf{74.3} & \textbf{6.2} \\
        \bottomrule[0.08em]
    \end{tabular}
    % }%
    \end{adjustbox}
    \vspace{-3mm}
    \caption{
        \textbf{Performance on Different Methods and Difficulties.} CC and UPC show Micro/Macro Pass Rate. Results are reported in percentage (\%). The best and second-best results are marked in bold and underlined.
    }
    \label{tab:main_result}
\end{table*}
\vspace{-3mm}

\paragraph{Incorrect Action Penalty}
To improve the reliability and robustness of the execution process, we introduce a penalty for invalid tool or memory invocations. An error is identified when \texttt{<tool\_response>} or \texttt{<memory\_response>} contains an explicit error message, upon which a penalty of $P_{\mathbf{action}}=-1$ is applied to the entire sequence, and $P_{\mathbf{action}}=0$ otherwise.

\paragraph{Travel Plan Reward}
To evaluate the quality of generated travel plans against both commonsense and user preference constraints, we design a two-stage gating reward. A format score $R_\mathbf{format} \in \{0,1\}$ first verifies structural correctness; only if passed, the Commonsense Constraint score $R_\mathbf{CC}$ and User Preference Constraint score $R_\mathbf{UPC}$ are assessed, ensuring the model learns to produce well-formed outputs before optimizing for content quality:
\begin{equation*}
R_{\mathbf{plan}} = R_{\mathbf{format}} \cdot \left( R_{\mathbf{CC}} + R_{\mathbf{UPC}} \right)
\end{equation*}
The total reward combines both components:
\begin{equation*}
R_\mathbf{total} =  P_\mathbf{action} +  R_\mathbf{plan}
\end{equation*}
This design penalizes any action-level failure while rewarding high-quality, preference-aligned plans, jointly encouraging reliable execution and personalized planning.

\subsection{Policy Optimization via GRPO} \label{sec:grpo}
Inspired by recent advances in RL-based LLM training \cite{tinyzero,kimiteam2025kimik15scalingreinforcement,wang2025actingreasoningmoreteaching,wang2025perceptionawarepolicyoptimizationmultimodal, docos}, we adopt GRPO \cite{shao2024deepseekmath} to optimize the policy model by maximizing the following objective:
\begin{align*}
& \mathcal{J}_{\mathrm{GRPO}}(\theta)=\mathbb{E}_{[\{y_{i}\}_{i=1}^{G}\sim\pi_{\theta}(Y|q)]}\frac{1}{G}\sum_{i=1}^{G}\frac{1}{|y_{i}|}\sum_{t=1}^{|y_{i}|}\Big\{ \\
 & \min\left[r_{i,t}(\theta)\hat{A}_{i,t},\mathrm{clip}\left(r_{i,t}(\theta),1-\epsilon,1+\epsilon\right)\hat{A}_{i,t}\right] \nonumber \\
 & -\beta\mathbb{D}_{KL}\left[\pi_{\theta}||\pi_{ref}\right]\biggr\} \nonumber \\
 & \nonumber  \mathrm{with} \quad r_{i,t}(\theta)=\frac{\pi_{\theta}(y_{i,t}|q,y_{i,<t})}{\pi_{\theta_{old}}(y_{i,t}|q,y_{i,<t})},
\end{align*}
where $\pi_\theta$ and $\pi_{ref}$ stand for the policy model and reference model respectively. $\mathbb{D}_{KL}$ is the KL-divergence, scaled by a coefficient $\beta$. $q$ denotes a query, and $y$ represents a sequence of structured actions.

\subsection{Training with Loss Mask} \label{sec:lossmask}
During training, the rollout contains externally sourced tokens within \texttt{<tool\_response>} and \texttt{<memory\_response>} that are not generated by the model. Including them in the loss computation introduces noise, so we follow \citet{jin2025searchr1trainingllmsreason} and apply a loss mask to exclude these segments from gradient updates, ensuring the model is optimized only on its own generated tokens.

\section{Experiments}
\subsection{Experimental Setup}
\paragraph{Baselines and Models}
Following prior works~\citep{xie2024travelplanner, toolspectrum}, we adopt two baselines: \textbf{ReAct} \cite{yao2023react} and \textbf{SFT}, alongside our proposed \textbf{B2T-Agent}, all with interactive reasoning where the model interacts with the environment to gather information. To ensure fair comparison, all methods are given the full user behavior trajectory. We evaluate open-source models including Qwen3-8/14/32B \cite{yang2025qwen3} and Deepseek-V3 \cite{liu2024deepseek}, as well as frontier proprietary models including GPT-4o and GPT-4.1.
\paragraph{Metrics}
Following \citet{xie2024travelplanner}, we report the Delivery Rate (\textbf{DR}), which measures the proportion of plans that successfully produce a structurally valid travel itinerary. For constraint adherence, we compute the Micro Pass Rate (\textbf{Micro PR}) and Macro Pass Rate (\textbf{Macro PR}) separately for Commonsense Constraints (\textbf{CC}) and User Preference Constraints (\textbf{UPC}), defined as:
\begin{equation*}
    \text{Micro PR} = \frac{\sum_{p \in P} \sum_{c \in C_p} \mathds{1}_{\text{passed}(c, p)}}{\sum_{p \in P} |C_p|}
\end{equation*}
\begin{equation*}
    \text{Macro PR} = \frac{\sum_{p \in P} \mathds{1}_{\text{passed}(C_p, p)}}{|P|}
\end{equation*}
For overall performance, we report the Final Pass Rate (\textbf{Final}), the percentage of plans satisfying all constraints, and the LLM Pass Rate (\textbf{LLM}) for flexible LLM-based evaluation.

\paragraph{Implementation Details}
We adapt the ReAct baseline from TravelPlanner~\cite{xie2024travelplanner} to our expanded toolset and sandbox constraints, and conduct SFT using LLaMAFactory~\cite{zheng2024llamafactory}. B2T-Agent is built on RLFactory~\citep{Simple-Efficient_RL-Factory_2025} with the prompt template shown in Figure~\ref{prompt-hta}.

\subsection{Main Results}
% 1. 现在的模型对于根据用户历史行为生成个性化的旅行计划的表现不好。可以从Table~\cite{tab:main_result}中看出，对于hard难度的即使是最好的闭源模型也仅仅能达到0.50的Final Pass Rate，且在简单难度的任务上，Final PR主要受Constraint Satisfaction 的影响，而到了Medium和Hard难度，由于用户个性化偏好维度和数量的增加，导致Final PR主要受User-Centric Alignment的影响。

% 2. 我们的方法B2T-Agent可以有效地解决当前任务。从Table~\cite{tab:main_result}中看出，B2T-Agent方法的大部分指标均为最优，仅用8B的模型即可超越ReAct方法的主流闭源模型；且相较于SFT方法而言，对于DR和User-Centric Alignment的提升更明显，这说明模型真正从自主与环境交互中学会了高效合理使用工具和记忆，且因人而异的生成符合约束的结果，并不是仅仅依靠对于固定模式的记忆，具体分析将会在后文的错误分析章节中详细讨论。

% 3. 从Table~\cite{tab:main_result}中看出，ReAct方法对模型本身的能力要求很高，才能有效的与环境交互最终生成完整的旅行计划，而且在约束的满足能力中，也是表现最差的；而经过SFT的模型，可以有效的从范式中学会与环境的交互方式，从而提升DR，而且对于每个query的CS约束都是相同的，模型也可以从已有数据中总结出规律，从而提升CS，但对于更灵活的UCA，需要根据每个用户特定的情况动态的调整生成的计划，仅依靠SFT提升效果不明显；而我们的方法B2T-Agent，依靠reward作为指导信号，激发模型的交互能力、并可以灵活的根据用户的不同需求生成更符合用户偏好的计划，对所有指标均有大幅提升。
Table~\ref{tab:main_result} reports the main results and we highlight three key findings.

\paragraph{\textbf{\textit{Existing models struggle with Action-Aware Travel Planning.}}} Even the strongest proprietary model, GPT-4.1, achieves only 0.5 Final PR on Hard tasks, underscoring the severity of the challenge. This stems from a fundamental shift in what drives success: while Easy tasks can largely be solved by satisfying CC, Medium and Hard tasks increasingly require alignment with UPC, a more personalized and dynamic requirement that current models consistently fail to capture. While LLM-based evaluation assigns relatively higher scores than rule-based Final due to its tolerance for output variations, both metrics confirm the same trend: UPC satisfaction remains the key bottleneck as task difficulty scales up.

\paragraph{\textbf{\textit{B2T-Agent achieves the best performance on Action-Aware Travel Planning.}}} B2T-Agent consistently achieves the best performance across all difficulty levels, with B2T-Agent-14B leading all baselines on both Final PR and LLM PR across Easy, Medium, and Hard splits. Furthermore, this advantage becomes especially pronounced on harder tasks: on the Hard split, B2T-Agent-14B achieves 5.0 Final PR compared to 0.5 for the best ReAct baseline (GPT-4.1), and 74.3 LLM PR compared to 28.2. Notably, even B2T-Agent with Qwen3-8B surpasses GPT-4.1 in average Final PR (4.7 vs. 3.2), highlighting the efficiency of our approach.

\paragraph{\textbf{\textit{B2T-Agent Achieves Superior Performance on User Preference Constraints.}}}
Since CC constraints are shared across all queries, SFT can learn fixed patterns to improve CC satisfaction; however, UPC constraints vary per user and require dynamic adaptation, which SFT fails to provide. In contrast, B2T-Agent leverages reward signals to drive the model toward personalized planning through interaction, rather than relying on memorized response patterns. This translates to substantial UPC gains: on the Hard split, B2T-Agent-14B achieves 5.0 UPC Macro PR, compared to 0.3 for SFT-14B and 0.8 for the best ReAct baseline (DeepSeek-V3).
\section{Analysis}
In this section, we present a comprehensive analysis aimed at addressing the following research questions. \textbf{RQ1:} \textit{Does incorporating behavior trajectory really help with personalized travel planning?} (\S\ref{sec: action_trajectory}) \textbf{RQ2:} \textit{How does increasing trajectory complexity affect method performance?} (\S\ref{sec:rq2}) \textbf{RQ3:} \textit{How does each component affect the performance of B2T-Agent?} (\S\ref{sec: ablation}) \textbf{RQ4:} \textit{Can B2T-Agent generalize to other travel planning benchmarks?} (\S\ref{sec: travelplanner})

\subsection{RQ1: Impact of Behavior Trajectory } \label{sec: action_trajectory}
\begin{figure}[!ht]
    \centering
    \includegraphics[width=\linewidth]{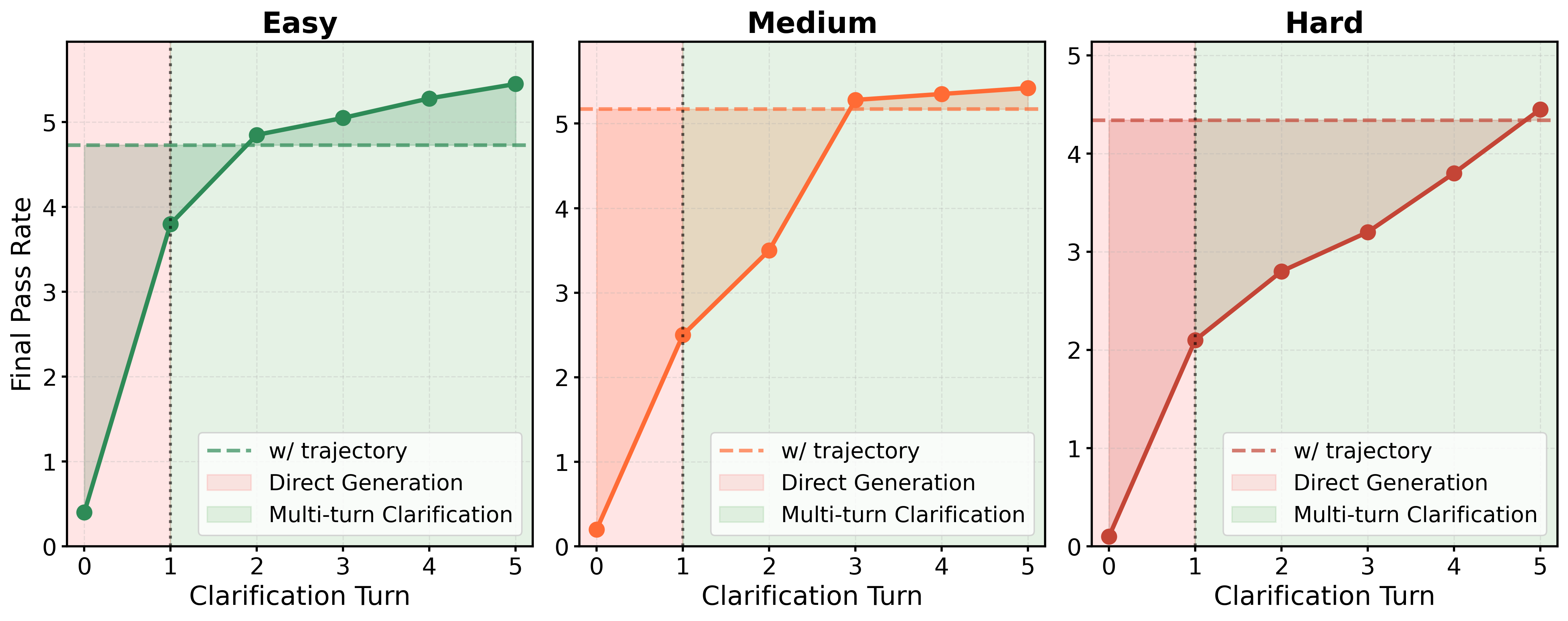}
    \caption{\textbf{Impact of Trajectory and Clarification Turns on Final PR.} Incorporating user trajectory significantly improves initial plan quality and reduces the need for clarification turns.}
    \label{fig:clarification}
\end{figure}

\noindent To investigate whether user behavior trajectory benefits personalized travel planning, we evaluate Qwen3-8B-B2T-Agent using only implicit user instructions, without trajectory. Instead, GPT-4.1 is used to simulate user feedback, allowing the model to iteratively refine plans through clarification turns. As shown in Figure~\ref{fig:clarification}, with zero clarification turns, Final PR is nearly zero, indicating failure to align plans with user preferences. To match the performance of trajectory-augmented planning, Easy, Medium, and Hard tasks require approximately 2, 3, and 5 clarification turns, respectively. These results confirm that user trajectory substantially improves initial plan quality and reduces reliance on multi-turn clarification.

\subsection{RQ2: Impact of Trajectory Complexity}\label{sec:rq2}
\begin{figure}[!ht]
    \centering
    \includegraphics[width=\linewidth]{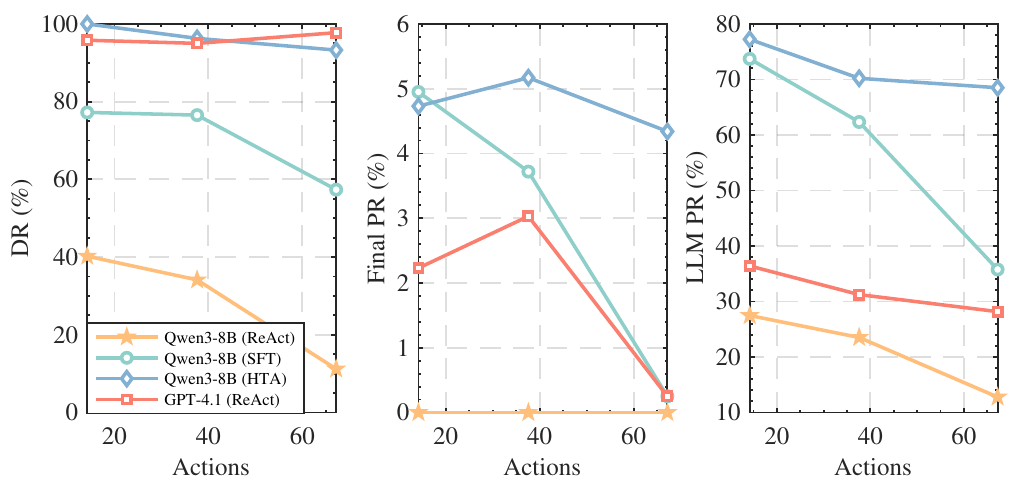}
    \caption{\textbf{Relationship between the number of actions in trajectories and the performance} of DR, Final PR, and LLM PR.}
    \label{fig:RQ1}
\end{figure}
We compare Qwen3-8B and GPT-4.1 across DR, Final PR, and LLM PR as trajectory complexity increases (measured by the number of actions). As shown in Figure~\ref{fig:RQ1}, Qwen3-8B under SFT and ReAct shows sharp drops across all metrics, as longer trajectories introduce more user preferences and edge cases beyond their capacity. In contrast, GPT-4.1 maintains strong DR but suffers notable declines in Final PR and LLM PR. Qwen3-8B-B2T-Agent performs best overall, sustaining high DR and stable PR scores; reward-driven interactive training enables it to handle edge cases and capture user preference–POI relationships that SFT and prompting cannot cover.

\subsection{RQ3: Ablation Study} \label{sec: ablation}
\definecolor{rowblue}{RGB}{235, 245, 255}
\begin{table}[!ht]
    \centering
    \resizebox{\columnwidth}{!}{
    \begin{tabular}{lccc}
        \toprule[0.08em]
        \textbf{Method} & \textbf{DR} & \textbf{Final PR} & \textbf{LLM PR}\\
        \midrule[0.05em]
        \rowcolor{rowblue}

        % 基线方法
        \textbf{B2T-Agent} & 96.52 & \textbf{4.75} & \textbf{71.98}\\
        % \midrule[0.03em]

        % 消融实验对比
        \textit{- w/o. M} & 
            \textbf{97.46} (\textcolor{mygreen}{$\uparrow$0.97\%}) & 
            4.21 (\textcolor{myred}{$\downarrow$11.37\%}) & 
            64.24 (\textcolor{myred}{$\downarrow$10.76\%}) \\

        \textit{- w/o. P} & 
            86.70 (\textcolor{myred}{$\downarrow$10.17\%}) & 
            4.32 (\textcolor{myred}{$\downarrow$9.05\%}) & 
            65.22 (\textcolor{myred}{$\downarrow$9.39\%}) \\

        \textit{- w/o. L} & 
            78.18 (\textcolor{myred}{$\downarrow$18.99\%}) & 
            3.79 (\textcolor{myred}{$\downarrow$20.21\%}) & 
            57.97 (\textcolor{myred}{$\downarrow$19.47\%}) \\
        
        \bottomrule[0.08em]
    \end{tabular}
    }
    \vspace{-3mm}
    \caption{\textbf{Ablation study on Qwen3-8B.}}\label{tab:ablation_results_selected}
    \vspace{-3mm}
\end{table}

\noindent We ablate each key component of B2T-Agent on Qwen3-8B to assess their individual contributions. As shown in Table~\ref{tab:ablation_results_selected}, removing the memory module (\textit{w/o. Memory}) slightly improves DR (+0.97\%), as the model skips memory operations and favors faster responses. However, without memory, the model loses access to user preferences, causing Final PR and LLM PR to drop by 11.4\% and 10.8\%, respectively. Without the penalty term (\textit{w/o. Penalty}), the model fails to correct inappropriate tool or memory actions, leading to repetitive loops and a 10.2\% drop in DR. Finally, removing loss masking (\textit{w/o. Loss Mask}) introduces noisy gradients from non-answer tokens, causing the largest overall decline, with DR, Final PR, and LLM PR dropping by 19.0\%, 20.2\%, and 19.5\%, respectively.

\subsection{RQ4: Cross-Benchmark Evaluation} \label{sec: travelplanner}
\definecolor{rowblue}{RGB}{235, 245, 255}
\begin{table}[!ht]
    \centering
    \resizebox{\columnwidth}{!}{
        \begin{tabular}{lcccccc}
            \toprule[0.08em]
            \textbf{Method} & \textbf{DR} & \textbf{CMi} & \textbf{CMa} & \textbf{HMi} & \textbf{HMa} & \textbf{FPR} \\
            \midrule[0.05em]
            
            Mistral-7B\textsuperscript{\textdagger} & 7.0 & 4.8 & 0.0 & 0.0 & 0.0 & 0.0 \\
            GPT-4-Turbo\textsuperscript{\textdagger} & 93.1 & 63.3 & 2.9 & 10.5 & 5.5 & 0.6 \\
            GPT-4.1 &  99.4&  75.1&  5.7&  10.7&  7.8& 1.5 \\
            \rowcolor{rowblue}
            \textbf{Qwen3-8B-B2T-Agent} & \textbf{100}& \textbf{76.8}& \textbf{25.0}& \textbf{26.5}& \textbf{10.2}& \textbf{9.0} \\
            \bottomrule[0.08em]
        \end{tabular}%
    }
    \vspace{-3mm}
    \caption{\textbf{Performance on the TravelPlanner} (\textsuperscript{\textdagger}results from the original paper).}
    \label{tab:travelplanner}
\end{table}
\noindent To validate the generality of B2T-Agent for travel planning, we further conduct a cross-benchmark evaluation on TravelPlanner~\cite{xie2024travelplanner}. As shown in Table~\ref{tab:travelplanner}, Qwen3-8B-B2T-Agent consistently outperforms all baselines, with especially large gains on constraint satisfaction (CMa: 25.0 vs. 5.7) and overall plan quality (FPR: 9.0 vs. 1.5). These results confirm that B2T-Agent generalizes effectively beyond our benchmark to standard travel planning settings.

\section{Conclusion}
In this paper, we identify a new task, Behavior-Aware Travel Planning, which infers user preferences from behavioral trajectories to generate personalized travel plans without requiring explicit input. To support this task, we present Behavior2Trip, a benchmark combining  user behavior data, and B2T-Agent, a reinforcement learning framework that enables models to interact autonomously using external tools and internal memory. Extensive experiments demonstrate that B2T-Agent effectively captures user intent, significantly reducing the need for explicit user interaction and enabling more seamless travel planning experiences.

\section*{Limitations}
While B2T-Agent achieves high-quality personalized planning by combining reward-driven multi-turn tool interaction with reinforcement learning, it incurs notable computational overhead from two main sources. First, during training, GRPO requires sampling multiple rollouts per query, which substantially increases computational cost. Second, at inference time, the multi-turn tool invocations lead to high first-token latency. Reducing this overhead without compromising planning quality remains an important direction for future work.

\section*{Ethics Statement}
POI and user-preference data can contain substantial sensitive information (e.g., consumers’ personally identifiable information), making it essential to protect user privacy from both the data and model perspectives. Specifically, as described in Section~\ref{sec: sandbox}, we first remove all explicit personally identifiable information from the POI database by using a locally deployed GPT-oss-120B~\citep{gpt_oss} to redact user reviews, business phone numbers, and POI descriptions. We then conduct random manual audits to verify that no privacy-sensitive content remains. In addition, during preference extraction, we model only high-level attributes rather than individual users, thereby further reducing privacy risk and mitigating potential bias when learning the mapping between preferences and POI categories.

\section*{Acknowledgments}

Thanks for the insightful comments and feedback from the reviewers. This work was supported by the National Natural Science Foundation of China (No. 62406015).

\bibliography{custom}

\appendix

\clearpage
\appendix

\section{Behavior2Trip}
\subsection{POI Database Details} \label{app:poi}
\begin{table*}[!ht]
\centering
\renewcommand{\arraystretch}{1.2}
\begin{tabularx}{\textwidth}{c c  X}
% --- 修改1：最上面的行变成 \toprule[0.08em] ---
\toprule[0.08em]
\textbf{POI Type} & \textbf{Count}  & \multicolumn{1}{c}{\textbf{Features}} \\
% --- 修改2：中间的行变成 \midrule[0.05em] ---
\midrule[0.05em]

Restaurant & 29,604  & \textit{Name, Details, Address, Phone, Province, City, Location, Longitude, Latitude, Tertiary Category, Secondary Category, Primary Category, Business Hours, Rating Rank, Parking Information, Average Price, Average Rating, User Reviews} \\
% --- 修改2：中间的行变成 \midrule[0.05em] ---
\midrule[0.05em]

Flights & 710,633  & \textit{Flight Number, Airline, Airline Code, Departure City, Arrival City, Departure City Code, Arrival City Code, Departure Airport, Arrival Airport, Departure Airport Code, Arrival Airport Code, Departure Date, Arrival Date, Departure Time, Arrival Time, Flight Duration, Departure Terminal, Arrival Terminal, Aircraft Model, Cabin Type, Cabin Code, Adult Fare, Child Fare, Adult Total Price, Child Total Price, Adult Fuel Surcharge, Adult Airport Construction Fee, Discount, On-time Rate, Meal Availability, Codeshare Flight} \\
\midrule[0.05em]

Hotels & 22,473  & \textit{Name, Details, Cover Image, Address, Phone, Province, City, District, Longitude, Latitude, Tertiary Category, Secondary Category, Primary Category, Business Hours, Rating Rank, Parking Information, WiFi Availability, Average Price, Average Rating, User Reviews, Room Information} \\
\midrule[0.05em]

Entertainment & 20,250  & \textit{Name, Details, Cover Image, Address, Phone, Province, City, District, Longitude, Latitude, Tertiary Category, Secondary Category, Primary Category, Business Hours, Rating Rank, Average Price, Average Rating} \\
\midrule[0.05em]

Trains & 10,371  & \textit{Departure City, Arrival City, Departure Station, Arrival Station, Train Number, Train Type, Travel Time, Travel Minutes, Departure Time, Arrival Time, Fare} \\
\midrule[0.05em]

Attractions & 8,229  & \textit{Name, Details, Cover Image, Address, Phone, Province, City, Location, Longitude, Latitude, Category Name, Attraction Type, Custom Type, Opening Hours, Best Visiting Season, Recommended Visiting Duration, Average Price, User Reviews, Ticket Information} \\
\midrule[0.05em]

Weather & 7,440  & \textit{City, Date, Max Temperature, Min Temperature, Weather, Wind Direction, Air Quality} \\
% --- 修改3：最下面的行变成 \bottomrule[0.08em] ---
\bottomrule[0.08em]
\end{tabularx}
\caption{Statistics of the collected \textbf{real-world} POI database. The dataset comprises \textbf{7} distinct categories totaling \textbf{809k} records, with an average of \textbf{17.7} attribute features per category.}
\label{tab:poi_info}
\end{table*}

Table~\ref{tab:poi_info} presents all the real Points of Interest (POIs) information used in this paper. It includes the POI types, the number of POIs of each kind, and the names of fields associated with each type. This comprehensive dataset provides a realistic and diverse sandbox environment for our research.

\subsection{Toolset Details} \label{app:toolset}

\begin{table*}[!ht]
\centering
\small % Use a smaller font size
\begin{tabular}{@{} l p{5cm} p{6.5cm} @{}}
\toprule[0.08em]
\textbf{Tool} & \textbf{Description} & \textbf{Parameter} \\
\midrule[0.05em]

AccommodationSearch & Accommodation options in a specified city are found, with support for preference-based filtering. &
\textbf{Main Parameters:}
\begin{itemize}[nosep,leftmargin=*,topsep=2pt]
    \item \texttt{city: str}
    \item \texttt{n: int (optional)}
    \item \texttt{preferences: dict (optional)}
\end{itemize}
\textbf{Preference Keys:}
\begin{itemize}[nosep,leftmargin=*,topsep=2pt]
    \item \texttt{star\_rating: list of str}
    \item \texttt{amenities: list of str}
\end{itemize} \\
\midrule[0.05em]

AttractionSearch & Attractions in a specified city are found, with support for preference-based filtering. &
\textbf{Main Parameters:}
\begin{itemize}[nosep,leftmargin=*,topsep=2pt]
    \item \texttt{city: str}
    \item \texttt{n: int (optional)}
    \item \texttt{preferences: dict (optional)}
\end{itemize}
\textbf{Preference Keys:}
\begin{itemize}[nosep,leftmargin=*,topsep=2pt]
    \item \texttt{attraction\_type: list of str}
    \item \texttt{walk\_tolerance: list of str}
\end{itemize} \\
\midrule[0.05em]

EntertainmentSearch & Entertainment options in a specified city are found, with support for preference-based filtering. &
\textbf{Main Parameters:}
\begin{itemize}[nosep,leftmargin=*,topsep=2pt]
    \item \texttt{city: str}
    \item \texttt{n: int (optional)}
    \item \texttt{preferences: dict (optional)}
\end{itemize}
\textbf{Preference Keys:}
\begin{itemize}[nosep,leftmargin=*,topsep=2pt]
    \item \texttt{type: list of str}
\end{itemize} \\
\midrule[0.05em]

FlightSearch & Flight information is retrieved, with support for preference-based filtering. &
\textbf{Main Parameters:}
\begin{itemize}[nosep,leftmargin=*,topsep=2pt]
    \item \texttt{origin: str}
    \item \texttt{destination: str}
    \item \texttt{departure\_date: str}
    \item \texttt{n: int (optional)}
    \item \texttt{preferences: dict (optional)}
\end{itemize}
\textbf{Preference Keys:}
\begin{itemize}[nosep,leftmargin=*,topsep=2pt]
    \item \texttt{cabin\_class: list of str}
\end{itemize} \\
\midrule[0.05em]

MapGuide & Navigation routes from a departure location to a destination are searched, including walking, driving, and public transit options, either within the same city or between different cities. &
\begin{itemize}[nosep,leftmargin=*,topsep=2pt]
    \item \texttt{origin\_cityName: str}
    \item \texttt{origin: str}
    \item \texttt{destination\_cityName: str}
    \item \texttt{destination: str}
\end{itemize} \\
\midrule[0.05em]

RestaurantSearch & Dining options in a specified city are explored, with support for preference-based filtering. &
\textbf{Main Parameters:}
\begin{itemize}[nosep,leftmargin=*,topsep=2pt]
    \item \texttt{city: str}
    \item \texttt{n: int (optional)}
    \item \texttt{preferences: dict (optional)}
\end{itemize}
\textbf{Preference Keys:}
\begin{itemize}[nosep,leftmargin=*,topsep=2pt]
    \item \texttt{cuisine: list of str}
    \item \texttt{price\_level: list of str}
    \item \texttt{environment: list of str}
\end{itemize} \\
\midrule[0.05em]

TrainSearch & Train information is retrieved. &
\begin{itemize}[nosep,leftmargin=*,topsep=2pt]
    \item \texttt{origin: str}
    \item \texttt{destination: str}
    \item \texttt{n: int (optional)}
\end{itemize} \\
\midrule[0.05em]

WeatherSearch & Weather information for a specified city and date is retrieved. &
\begin{itemize}[nosep,leftmargin=*,topsep=2pt]
    \item \texttt{city: str}
    \item \texttt{date: str (optional)}
    \item \texttt{n: int (optional)}
\end{itemize} \\
\midrule[0.05em]

Python & Python code is executed, with the ability to display outputs and handle errors. &
\begin{itemize}[nosep,leftmargin=*,topsep=2pt]
    \item \texttt{code: str}
\end{itemize} \\

\bottomrule[0.08em]
\end{tabular}
\caption{The toolset definition comprising 9 specialized APIs. Each retrieval tool is equipped with fine-grained parameters and preference keys (e.g., \textit{cuisine}, \textit{star\_rating}) to support precise, multi-criteria filtering of the POI database.}
\label{tab:toolset}
\end{table*}

Table~\ref{tab:toolset} presents all the tools, including the tool names, descriptions, and their respective parameters. A total of nine tools are provided, each supporting fine-grained combinations based on user preference information, enabling precise retrieval of the most relevant data from the large POI database.

\subsection{Constraint Details} \label{app:constraint}
\begin{table*}[!ht]
\centering
\small
\begin{tabularx}{\textwidth}{l l X}
\toprule[0.08em]
\textbf{Category} & \textbf{Constraint} & \multicolumn{1}{c}{\textbf{Description}} \\
\midrule[0.05em]

Format (1) & JSON Format Validation & Validates overall JSON structure and required fields \\
\midrule[0.05em]

\multirow{4}{*}{Authenticity (4)} & POI Authenticity & Verifies that Points of Interest exist in real city databases \\
& Major Transport Authenticity & Confirms availability of intercity transport like flights/trains \\
& Local Transport Authenticity & Ensures local transport durations are reasonable \\
& Business Hours Authenticity & Checks that POI visits occur within operating hours \\
\midrule[0.05em]

\multirow{15}{*}{Reasonableness (15)} & Complete Meal Arrangement & Ensures three meals are arranged each day \\
& Non-duplicate Restaurants & No repeated restaurant recommendations across the plan \\
& Reasonable Meal Times & Meals occur at appropriate times of day \\
& Daily Accommodation Arrangement & A hotel must be arranged for each night \\
& Reasonable Accommodation Time & Hotel check-in should be scheduled after 18:00 \\
& Avoid Midnight Travel & No transport activities between 23:00 and 06:00 \\
& No Attractions After Late Major Transport & No sightseeing after arriving at a city after 18:00 \\
& Local Transport Time Limit & Total daily local transport should not exceed 2 hours \\
& Weather and City Info Included & Weather and city background must be included for each destination \\
& At Least One Attraction & Each day must feature at least one valid attraction \\
& Attraction Quantity Limit & No more than 4 attractions per day \\
& No Time Conflicts & Activities should not overlap, with at least 30-minute gaps \\
& Reasonable Time Range & Activities must occur between 09:00 and 21:00 \\
& Inter-attraction Transport Time & Transport between attractions should not exceed 1 hour \\
& No Long Continuous Gaps & No idle gaps longer than 2 hours during 09:00–21:00 \\
\midrule[0.05em]

\multirow{10}{*}{Personalization (10)} & Hotel Star Level Matching & Hotel meets user's minimum star rating \\
& Hotel Amenities Matching & Hotel has required facilities (e.g. WiFi, parking) \\
& Attraction Type Matching & Attractions align with user's preferences \\
& Walking Tolerance Matching & Plan respects walking endurance limitations \\
& Cuisine Type Matching & Meals reflect preferred cuisine types \\
& Price Level Matching & All items stay within specified price levels \\
& Dining Environment Matching & Dining settings match user preferences (e.g. quiet, lively) \\
& Transport Mode Matching & Intercity transport matches preferred modes \\
& Seat Class Matching & Matches user's preferred flight/train seat class \\
& Budget Constraint & Total cost stays within user-defined budget \\
\midrule[0.05em]

\multirow{2}{*}{Information Richness (2)} & Information Sufficiency & Each activity includes all required fields \\
& Information Richness & Extra details (e.g. descriptions, tips) are included when possible \\
\bottomrule[0.08em]
\end{tabularx}
\caption{Taxonomy of constraints derived from \textbf{real-world business scenarios}. The system enforces a total of \textbf{32} distinct constraints spanning \textbf{5} categories (e.g., Authenticity, Reasonableness) to ensure the feasibility and quality of travel plans.}
\label{tab:constraints}
\end{table*}

Table~\ref{tab:constraints} lists the constraints used to evaluate the quality of travel plans. A total of 32 constraints are categorized into five groups: Format (1), Authenticity (4), Reasonableness (15), Personalization (10), and Information Richness (2). Each constraint is derived from real-world business scenarios, providing a realistic and effective basis for assessing the quality of a travel plan.

\subsection{Preference Details} \label{app:preference}
\begin{table*}[!ht]
\centering
\renewcommand{\arraystretch}{1.3}
\begin{tabularx}{\textwidth}{l l >{\raggedright\arraybackslash}X}
\toprule[0.08em]
\multicolumn{1}{c}{\textbf{Field}} & 
\multicolumn{1}{c}{\textbf{Data Type}} & 
\multicolumn{1}{c}{\textbf{Value}} \\
\midrule[0.05em]

\multicolumn{3}{l}{\textbf{Historical Travel Experiences}} \\
accommodations & array (of object) & - \\
restaurants & array (of object) & - \\
attractions & array (of object) & - \\
transportation & array (of object) & - \\
\midrule[0.05em]

\multicolumn{3}{l}{\textbf{Hotel Preferences}} \\
star\_requirement & array (of string) & \textit{2-Star \& Below/Economy, 4-Star/Upscale, 3-Star/Comfort, 5-Star/Luxury, City Homestay} \\
essential\_facilities & array (of string) & \textit{Free WiFi, Private Bathroom, Parking, Gym, Swimming Pool, Restaurant} \\
\midrule[0.05em]

\multicolumn{3}{l}{\textbf{Attraction Preferences}} \\
preferred\_attraction\_types & array (of string) & \textit{Park/Square, Museum/Exhibition, Cultural/Historic Site, Cityscape, Natural Landscape, Comprehensive Scenic Area, Religious Site, Theme Park, Outdoor Adventure, Water Activities, Hot Spring/Wellness, Rural/Folk Village, Animal Watching, Ice/Snow Sports, Water Conservancy Facility} \\
walking\_tolerance & string & \textit{Low (1-3km), Medium (3-5km), High (5km+)} \\
\midrule[0.05em]

\multicolumn{3}{l}{\textbf{Dining Requirements}} \\
cuisine\_preferences & array (of string) & \textit{Beverages, Snacks/Fast Food, Other Cuisines, Hot Pot, Western Food, Cantonese, Other Chinese Food, Shanghai/Zhejiang Cuisine, Sichuan Cuisine, Bakery/Desserts, Barbecue, Buffet, Northern Chinese Food, Hunan Cuisine, Seafood, Japanese/Korean Cuisine, Southeast Asian Cuisine} \\
price\_level & string & \textit{Budget (30-80 CNY/person), Moderate (80-150 CNY/person), Generous (150-300 CNY/person), Luxury (300-500 CNY/person)} \\
dining\_environment & string & \textit{Quiet, Scenic View, Family-friendly, Lively, Unique Decor} \\
\midrule[0.05em]

\multicolumn{3}{l}{\textbf{Transportation Preferences}} \\
intercity\_transport & string & \textit{High-speed Rail, Regular Train, Airplane, Long-distance Bus} \\
intracity\_transport & string & \textit{Subway, Bus, Taxi, Ride-sharing, Walking} \\
seat\_class & string & \textit{Economy Class, Business Class, First Class Seat, Second Class Seat} \\
\bottomrule[0.08em]
\end{tabularx}
\caption{The schema of the user preference profile derived from \textbf{real-world travel data}. It consists of \textbf{14} fine-grained attributes across \textbf{5} dimensions (e.g., Dining, Transportation), defining both historical records and specific constraints to enable personalized planning.}
\label{tab:preference}
\end{table*}

Table~\ref{tab:preference} presents the fields and value ranges related to travel preferences defined in user profiles, including Historical Travel Experiences, Hotel Preferences, Attraction Preferences, Dining Requirements, and Transportation Preferences. Each field is closely aligned with POIs in the database, ensuring that user instructions with diverse preferences can be effectively addressed.

\section{Additional  Experimentss}

% \subsection{Main Results}
% \input{table/main_result_mono}
% Due to space limitations in the main text, we only reported the Macro Pass Rate for the CC and UPC constraints. Here, we provide the Micro Pass Rate as a fine-grained complement to the results shown in Table~\ref{tab:main_result}.

\subsection{RQ5: Case Study} \label{app:case}
\begin{figure*}[!htbp]
    \centering
    \includegraphics[width=\textwidth]{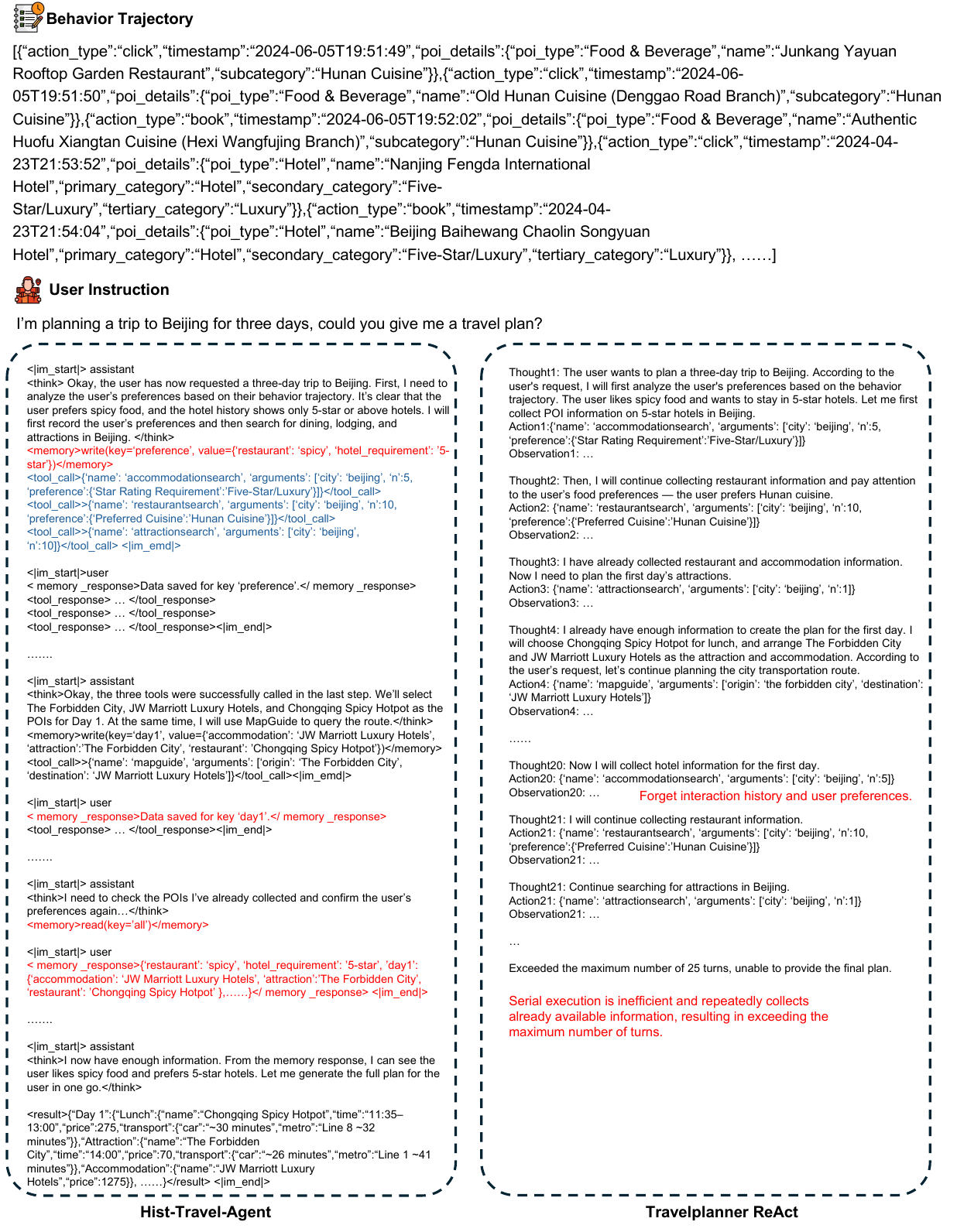}
    \caption{Case study comparing the reasoning trajectories of TravelPlanner ReAct and B2T-Agent. While the ReAct baseline suffers from context loss and redundant tool usage due to strictly sequential execution, B2T-Agent utilizes parallel tool invocation and a memory module to retrieve user preferences, ensuring efficient and accurate plan generation.}
    \label{fig:case_study}
\end{figure*}
As illustrated in Figure~\ref{fig:case_study}, we compare the performance of B2T-Agent and TravelPlanner ReAct when processing the same input. The ReAct model performs tool invocation in a strictly sequential manner, relying entirely on prompts to execute actions step by step. As the number of turns increases, the dialogue context grows rapidly, making it difficult for the model to retain previous interaction history and user preferences. This often leads to repeated tool usage and a lack of filtering based on user preferences. Such inefficiency not only increases the number of interaction turns but may also prevent the generation of a final travel plan due to the maximum turn limit.

In contrast, B2T-Agent supports parallel tool invocation and incorporates a memory module to read and write critical user information. During the planning process, the model proactively accesses stored user preferences and interaction history, thereby improving tool invocation efficiency and generating travel plans that better align with user needs.

\subsection{RQ6: Error Analysis} \label{app:error}
We perform a detailed error analysis on the two best‐performing models from our main experiment and our proposed B2T-Agent. From each model’s output, we randomly sampled 30 generated trajectories and manually examined them (total of 180 samples), identifying and categorizing five classes of recurring errors.
% 1. 模型仅以生成格式标准的旅行计划为目标，而忽略了约束的限制
% 2. 因为错误的工具或记忆调用导致一直死循环
% 3. 因为模型幻觉导致生成的旅行计划中包含虚构poi
% 4. 模型的上下文能力限制
% 5. 模型生成的旅行计划与要求格式不符，导致无法完成后续解析评估工作
\subsubsection{\textit{Constraint Violations}}
% 模型在输出旅行计划时，只考虑了是否符合格式、完整性等要求，而忽视了推荐poi与现实约束的联系，也不考虑用户的偏好，导致生成的绝大部分规划都不能直接供用户使用
Agents often generate travel plans that ignore real-world constraints, such as geographic feasibility, time budgets, and user preferences. While the outputs usually satisfy format and completeness requirements, they fail to consider whether the recommended POIs are realistically visitable or aligned with user needs. As a result, most plans are not directly usable.
\subsubsection{\textit{Erroneous Tool or Memory Loops}}
% 在逐步与外部环境或内部记忆交互的过程中，若生成的tool call工具格式或参数错误，则会返回错误，模型不会根据错误信息改正自己的调用方式，从而陷入死循环，导致生成过程崩溃。
When interacting step-by-step with external tools or internal memory, the models sometimes generate invalid tool call formats or incorrect parameters. Once an error occurs, the model does not revise its behavior based on feedback, leading to repeated failures and ultimately crashing the generation process.
\subsubsection{\textit{Hallucinated POIs}}
% 模型对内部知识边界认识有限，导致幻觉，即在生成旅行计划时表现得很明显，生成的计划中包含不存在database中的POI，而是使用保存在参数中错误的poi。
Agents occasionally include POIs that do not exist in the database. This type of hallucination indicates a poor understanding of the model's knowledge boundaries. Instead of retrieving POIs from verified data, the model relies on memorized or incorrect internal representations.

\subsubsection{\textit{Contextual Reasoning Limits}}
% 在生成旅行计划的过程中，模型会与环境和记忆中交互从而得到大量的信息，会导致上下文的大量堆积，每一个轨迹都超过20ktoken，模型会因为上下文的能力太弱，导致不能从交互的轨迹中提取关键信息，从而影响生成的计划质量。
Each generation involves extensive interaction with memory and environment, leading to long contextual sequences, often exceeding 20k tokens. The models struggle to extract relevant signals from such complex contexts, which negatively impacts their ability to generate coherent and relevant plans.

\subsubsection{\textit{Formatting Errors}}
% 模型的指令遵循能力有限导致模型不能输出我们规定的旅行计划格式，从而不能满足本文评估以及真实场景中应用的需要。
Despite clear instruction prompts, the models sometimes fail to follow the required format for travel plans. These formatting issues hinder both automatic evaluation and real-world usability, revealing limitations in instruction following and output consistency.
% \subsection{Hyperparameter Details}
% We present the hyperparameters used for training the B2T-Agent model within the RL Factory framework.

% \begin{itemize}
%     \item \textbf{Train Batchsize}: 128
%     \item \textbf{Max Prompt Length}: 28000
%     \item \textbf{Max Response Length}: 2000
%     \item \textbf{Mini Batchsize Per GPU}: 1
%     \item \textbf{Micro Batchsize Per GPU}: 1
%     \item \textbf{Rollout Number}: 8
%     \item \textbf{Max Turn}: 15
%     \item \textbf{KL Coefficient}: $1 \times 10^{-3}$
%     \item \textbf{Learning Rate}: $1 \times 10^{-6}$
%     \item \textbf{Epoch}: 1
% \end{itemize}

\section{Prompt Template}
In this section, we present the prompts used in this study. Figures~\ref{prompt-next-step-prompt}, \ref{prompt-profile}, \ref{prompt-trajectory}, \ref{prompt-quality-control}, and \ref{prompt-hta} illustrate the \textbf{Next Step Prompt}, \textbf{Profile Filtering}, \textbf{Behavior Trajectory Generation}, \textbf{Quality Control}, and \textbf{B2T-Agent}, respectively.

\begin{figure*}[!h]
\centering
\begin{mybox}{Prompt Template for Next Step Prompt}
\begin{lstlisting}[style=promptstyle]
Review the user's goals, summarize the information already collected from past conversations, and then briefly think about what additional information is needed to create a travel guide as detailed as the template, before selecting multiple tools to execute.
Notes:
Do not aim too high. You have multi-turn problem-solving ability, so to ensure the completeness and accuracy of your output, you can solve only part of the problem each time.
Keep your reasoning concise: output your reasoning in 
 with no more than 100 words, focusing only on the most important information and avoiding repetitive details.
You must use all the information obtained through tools to summarize your answer. In the 
 section, do not guess dates, destination city attractions, or attraction details.
In the first reasoning step, you can only call 3 tools to determine three pieces of information, and only after that can you execute other operations.
If the city is different from the user's current location, you need to check for round-trip flights/trains. If the user hasn't specified a departure city, assume departure is from the local city, in which case no train/flight information is needed.
You must end the task using <result> </result>. The <result> </result> tag must stand alone and cannot be combined with other tool calls.
\end{lstlisting}
\end{mybox}
\caption{Prompt Template for Next Step Prompt.}
\label{prompt-next-step-prompt}
\end{figure*}

\begin{figure*}[ht]
\begin{mybox}{Prompt Template for Filtering Profile}
\begin{lstlisting}[style=promptstyle]
Evaluate whether the following persona is reasonable:
Age: (*\textcolor{blue}{\{age\}}*)
Education Background: (*\textcolor{blue}{\{education\}}*)
Occupation: (*\textcolor{blue}{\{occupation\}}*)
Location: (*\textcolor{blue}{\{location\}}*)
Family Status: (*\textcolor{blue}{\{family\_status\}}*)
Sightseeing Preferences: (*\textcolor{blue}{\{sightseeing\_preference\}}*)
Food Preferences: (*\textcolor{blue}{\{food\_preference\}}*)
Hotel Preferences: (*\textcolor{blue}{\{hotel\_preference\}}*)
Transportation Preferences: (*\textcolor{blue}{\{transportation\_preference\}}*)

Evaluation criteria include but are not limited to:
1. Whether the age matches the education level (e.g., a 20-year-old PhD is unlikely; a 40-year-old with only elementary education is rare)
2. Whether the education background aligns with the occupation (e.g., someone with only a primary school education being a surgeon is unreasonable; a master's degree holder in a research or senior management role makes more sense)
3. Whether the age aligns with the family status (e.g., an 18-year-old being married with children is uncommon; a 60-year-old single person with no children may require justification)
4. Whether the occupation logically fits the location (e.g., a tech product manager is usually based in major cities; a full-time fashion influencer in a rural town would need context)
5. Whether the income level inferred from the occupation matches the stated preferences (e.g., someone earning a low income preferring luxury hotels and Michelin restaurants is inconsistent)
6. Whether the transportation preferences match the age or financial condition (e.g., a 70-year-old backpacking solo on a tight budget is uncommon; older people tend to prefer comfort and convenience)
7. Whether sightseeing/food/hotel preferences align with the overall persona (e.g., a highly educated researcher preferring trendy Instagram spots may seem off; a low-income person favoring ultra-luxury hotels may be unrealistic)
8. Whether the overall persona is logically coherent (e.g., a culturally inclined person should show that interest in their dining, travel, or accommodation preferences)

Please respond strictly in the following JSON format and do not add any extra content:
{"reasonable": true/false, "reason": "specific reason"}

Output: (*\textcolor{blue}{\{output\}}*)
\end{lstlisting}
\end{mybox}
\caption{Prompt Template for Filtering Profile.}
\label{prompt-profile}
\end{figure*}

\begin{figure*}[!htbp]
\centering
\begin{mybox}{Prompt Template for Behavior Trajectory Generation}
\begin{lstlisting}[style=promptstyle]
# ROLE
You are role-playing a real user on a Chinese online travel platform (e.g.,
Ctrip / Fliggy). Your task is to generate a single-session behavior trajectory
that authentically reflects the given user profile, as if you were genuinely
browsing and exploring the platform during one visit.

# USER PROFILE
(*\textcolor{blue}{\{user\_profile\}}*)

# SESSION CONTEXT
Current session number: (*\textcolor{blue}{\{session\_id\}}*)
Previous session history (for cross-session continuity):
(*\textcolor{blue}{\{previous\_sessions\}}*)

# CANDIDATE POI POOL
## Preference-Matched POIs (aligned with the user's profile):
(*\textcolor{blue}{\{matched\_pois\}}*)
## Noise POIs (NOT aligned with the user's profile):
(*\textcolor{blue}{\{noise\_pois\}}*)

# ACTION SPACE
- "click"            : The user is browsing or curious. Low commitment.
- "add_to_favorites" : The user finds the POI appealing and saves it for later.
- "book"             : The user commits to the POI.

# BEHAVIORAL RULES
1. Preference Grounding: "add_to_favorites" and "book" MUST use Preference-Matched
   POIs only. Noise POIs MUST only receive "click" actions.
2. Behavioral Funnel: click -> add_to_favorites -> book. A POI must be clicked
   before it is favorited; favorited before it is booked. Most POIs stop at click.
3. Sequential Coherence:
   - Within session: consecutive actions should be topically related.
   - Cross-session: if previous sessions are provided, continue the established
     preference trajectory; do not re-book already-booked POIs; early sessions
     reflect broad exploration, later sessions narrow toward commitment.
4. Spatio-Temporal Consistency: POIs in one session should be geographically
   plausible. Decide transportation and accommodation before attractions and dining.
5. Semantic / Category Diversity: cover multiple POI categories per trajectory.
6. Noise Behavior: noise clicks should be sporadically interleaved, not clustered.
7. Realism Constraints: do NOT invent POIs outside the candidate pool. Do NOT
   mention preferences explicitly. Do NOT include reasoning or commentary.

# OUTPUT FORMAT
Return a JSON array for THIS SESSION ONLY. Each record:
[
  {
    "action_type": "click" | "add_to_favorites" | "book",
    "poi_id": "<string from candidate pool>",
    "poi_name": "<string from candidate pool>"
  },
  ...
]
Timestamps and full POI details are added automatically in post-processing.
\end{lstlisting}
\end{mybox}
\caption{Prompt Template for Behavior Trajectory Generation.}
\label{prompt-trajectory}
\end{figure*}

\begin{figure*}[!htbp]
    \centering
    \begin{mybox}{Prompt Template for Quality Control}
\begin{lstlisting}[style=promptstyle]
You are a professional travel-user behavior data-quality analyst.

Your task is to evaluate the given user data. You must imitate the example below, score each of the three dimensions--'Completeness of Trajectory Information', 'Clarity of Preference Reflection in the Trajectory', and 'Degree to Which the Plan Meets Constraints'--with either 0 or 1 point, and provide a structured JSON output. The output should include the scores, your reasoning, and specific instructions that can be used to automate data modifications.

Scoring Criteria
1. Completeness of Trajectory Information
   - 1 point (Meets): The trajectory contains a clear behavior chain from browsing/exploration to final decision (e.g., multiple behaviors such as clicks, favorites, bookings/orders) and ends with a definitive decision point.
   - 0 points (Does Not Meet): The trajectory is overly single-patterned (e.g., only a few clicks), lacks a final decision step, or the behavior chain is broken/illogical.

2. Clarity of Preference Reflection in the Trajectory
   - 1 point (Meets): The user's key behaviors (such as favoriting, booking) are highly consistent with the preferences stated in persona scenic spot preference.
   - 0 points (Does Not Meet): The user's key behaviors conflict with or are completely unrelated to the stated preferences.

3. Degree to Which the Plan Meets Constraints
   - 1 point (Meets): The generated travel plan satisfies all constraints.
   - 0 points (Does Not Meet): The generated travel plan clearly conflicts with or is completely unrelated to the constraints.

Output Format
You must strictly output according to the JSON structure shown in the example and do not add any explanatory text outside the JSON object.

Predefined Operation-Type Codes
- No action required: Data quality is high; no modification is needed.
- Correct user preference: When behavior conflicts with declared preferences, adjust the preferences in the persona based on behavioral history.
- Mark incomplete trajectory: When the trajectory lacks a final decision step, mark the data so it can be down-weighted or filtered in downstream tasks.
- Mark for deletion: When data quality is too low to fix, recommend deleting the entire record.

Input: (*\textcolor{blue}{\{input\}}*)

Output: (*\textcolor{blue}{\{output\}}*)
\end{lstlisting}
\end{mybox}
\caption{Prompt Template for Quality Control.}
\label{prompt-quality-control}
\end{figure*}

\begin{figure*}[!htbp]
\centering
\begin{mybox}{Prompt Template for B2T-Agent}
\begin{lstlisting}[style=promptstyle]
You are a seasoned travel planner with extensive experience in generating personalized plans.
Your core task is to deeply understand the user's requirements, gather information by calling tools, and ultimately design a thorough and meticulous travel plan.
You possess an internal dictionary-structured memory module for storing and retrieving key information. Use <memory></memory> to wrap any operation that reads or writes session memory using "action": "read" or "write" with a "key" (and optional "value" for write). After gathering all necessary information and crafting the complete plan, please wrap the final itinerary in JSON format between the "<answer>" and "</answer>" tags.
Note: The delivered travel plan must be complete, covering every day of the trip.
Please note that the final output should be a JSON list (array) that includes all travel days.
Each day is an independent JSON object that follows the same internal structure.

# Tools
{"name": "AccommodationSearch", "parameters": {"type": "object", "properties": {"city": {"title": "City", "type": "string"}, "n": {"default": 5, "title": "N", "type": "integer"}, "preferences": {"additionalProperties": true, "default": null, "title": "Preferences", "type": "object"}}, "required": ["city"]}}
{"name": "RestaurantSearch", "parameters": {"type": "object", "properties": {"city": {"title": "City", "type": "string"}, "n": {"default": 5, "title": "N", "type": "integer"}, "preferences": {"additionalProperties": true, "default": null, "title": "Preferences", "type": "object"}}, "required": ["city"]}}
{"name": "FlightSearch", "parameters": {"type": "object", "properties": {"origin": {"title": "Origin", "type": "string"}, "destination": {"title": "Destination", "type": "string"}, "departure_date": {"title": "Departure Date", "type": "string"}, "n": {"default": 5, "title": "N", "type": "integer"}, "preferences": {"additionalProperties": true, "default": null, "title": "Preferences", "type": "object"}}, "required": ["origin", "destination", "departure_date"]}}
{"name": "TrainSearch", "parameters": {"type": "object", "properties": {"origin": {"title": "Origin", "type": "string"}, "destination": {"title": "Destination", "type": "string"}, "n": {"default": 5, "title": "N", "type": "integer"}}, "required": ["origin", "destination"]}}
{"name": "WeatherSearch", "parameters": {"type": "object", "properties": {"city": {"title": "City", "type": "string"}, "date": {"default": null, "title": "Date", "type": "string"}, "n": {"default": 5, "title": "N", "type": "integer"}}, "required": ["city"]}}
{"name": "EntertainmentSearch", "parameters": {"type": "object", "properties": {"city": {"title": "City", "type": "string"}, "n": {"default": 5, "title": "N", "type": "integer"}, "preferences": {"additionalProperties": true, "default": null, "title": "Preferences", "type": "object"}}, "required": ["city"]}}
{"name": "AttractionSearch", "parameters": {"type": "object", "properties": {"city": {"title": "City", "type": "string"}, "n": {"default": 5, "title": "N", "type": "integer"}, "preferences": {"additionalProperties": true, "default": null, "title": "Preferences", "type": "object"}}, "required": ["city"]}}
{"name": "MapGuide", "parameters": {"type": "object", "properties": {"origin_cityName": {"title": "Origin Cityname", "type": "string"}, "origin": {"title": "Origin", "type": "string"}, "destination_cityName": {"title": "destination_cityname", "type": "string"}, "destination": {"title": "Destination", "type": "string"}}, "required": ["origin_cityName", "origin", "destination_cityName", "destination"]}}

Behavior trajectory: (*\textcolor{blue}{\{behavior trajectory\}}*)
Instruction: (*\textcolor{blue}{\{instruction\}}*)
\end{lstlisting}
\end{mybox}
\caption{Prompt Template for B2T-Agent.}
\label{prompt-hta}
\end{figure*}

\end{document}